\documentclass{article}

\usepackage[final,nonatbib]{neurips_2023}

\usepackage[utf8]{inputenc} 
\usepackage[T1]{fontenc}    
\usepackage{hyperref}       
\usepackage{url}            
\usepackage{booktabs}       
\usepackage{amsfonts}       
\usepackage{nicefrac}       
\usepackage{microtype}      
\usepackage{xcolor}         
\usepackage{graphicx}
\usepackage{longtable}
\usepackage[numbers]{natbib}

\title{ NLLB-CLIP -- train performant multilingual image retrieval model on a budget }

\author{%
  Alexander Visheratin \\
  \texttt{alexvish91@gmail.com} \\
}

\begin{document}

\maketitle

\begin{abstract}
  Today, the exponential rise of large models developed by academic and industrial institutions with the help of massive computing resources raises the question of whether someone without access to such resources can make a valuable scientific contribution. To explore this, we tried to solve the challenging task of multilingual image retrieval having a limited budget of \$1,000. As a result, we present NLLB-CLIP -- CLIP model with a text encoder from the NLLB model. To train the model, we used an automatically created dataset of 106,246 good-quality images with captions in 201 languages derived from the LAION COCO dataset. We trained multiple models using image and text encoders of various sizes and kept different parts of the model frozen during the training. We thoroughly analyzed the trained models using existing evaluation datasets and newly created XTD200 and Flickr30k-200 datasets. We show that NLLB-CLIP is comparable in quality to state-of-the-art models and significantly outperforms them on low-resource languages.
\end{abstract}

\section{Introduction and model description}
\label{sec:nllb-clip}

Contrastive Language-Image Pre-Training (CLIP) \cite{radford2021learning} is a powerful architecture that allows achieving high-quality results on a variety of tasks, such as zero-shot classification, text-image, and image-text extraction. It uses vision and text transformers \cite{dosovitskiy2020image, vaswani2017attention} to encode information from images and texts into the common latent space. CLIP can be applied to new tasks without any fine-tuning \cite{thengane2022clip, li2022adapting} or be extended to solve more complex tasks as semantic segmentation \cite{lin2023clip}.

CLIP has been adapted to languages other than English, like Italian \cite{bianchi2021contrastive} and Chinese \cite{yang2022chinese}. Numerous works also demonstrated that CLIP can be extended to multiple languages at once \cite{carlsson2022cross, chen2022altclip}. But to date, there have been no capable CLIP-like models for low-resource languages \cite{nekoto2020participatory,joshi2020state}.

Recently, the "No Language Left Behind" (NLLB) model \cite{costa2022no} was introduced to enable translation between more than 200 languages. The model shows great performance in both high- and low-resource languages. Notably, the model has encoder-decoder architecture and was trained in different sizes (600M, 1.3B, and 3.3B parameters) that allow a wider variety of deployments.

The core question we investigated is whether we can use a pre-trained text encoder from NLLB models to extend CLIP capabilities to the languages of the Flores-200 dataset and stay within a limited budget of \$1,000. To answer this, we replaced the standard text encoder of the OpenAI CLIP \cite{radford2021learning} with the text encoder from the NLLB model. We left the rest of the model and loss functions the same as we wanted to understand the impact of this specific change.

For the experiments, we used various backbone models. For the image encoder, we used the original CLIP ViT base (denoted as "b" in experiments) and large ("l") and CLIP ViT huge ("h") trained by LAION. For the text encoder, we used respective parts of three NLLB variants -- base ("b"), large ("l"), and huge ("h").

\section{Datasets}
\subsection{Training dataset}

Since there are no image-text datasets available for all Flores-200 languages, we had to create a new, sufficiently large dataset to train the model. We used a random subset of the LAION COCO dataset\footnote{\url{https://laion.ai/blog/laion-coco/}} that contains automatically generated captions in MS COCO \cite{lin2014microsoft} style. We used the LAION aesthetic predictor model\footnote{\url{https://laion.ai/blog/laion-aesthetics/}} to filter the images during the processing and preserved only the pictures with an aesthetic score higher than 4.5. The threshold score was obtained empirically by manually analyzing 500 scored images from the dataset. We aimed to collect enough data but not too much so that we would stay within our budget when training the models. As a result, we got 106,246 images. English captions were translated into 200 languages of the Flores-200 dataset using the NLLB-3.3B model. We left 15\% of the dataset (15,937 samples) for validation. The result is the LAION-COCO-NLLB dataset \cite{laion-coco-nllb}, which we make publicly available. To the best of our knowledge, this is the largest image-text dataset in terms of languages covered. The key property of this dataset is that all 201 languages are presented equally, which greatly affects the model performance for low-resource languages, as shown in Section \ref{sec:evaluation}.

\subsection{Evaluation datasets}
\label{sec:eval-datasets}

We used two existing evaluation datasets for the experiments. (1) XTD10 \cite{aggarwal2020zeroshot} -- 1,000 image-text pairs from COCO 2014 dataset translated into 10 languages. It extends previous works \cite{rajendran-etal-2016-bridge, yoshikawa-etal-2017-stair} with 7 new languages. (2) Crossmodal-3600 \cite{ThapliyalCrossmodal2022} -- 3,600 images annotated with captions in 36 languages. The main advantage of this dataset is that it covers many languages, including five low-resource ones -- Bengali, Cusco Quechua, Maori, Swahili, and Telugu. 

To test the performance of the models using all Flores-200 languages, we created two new datasets. (1) XTD200 -- 1,000 English captions from XTD10 dataset translated to 200 languages using NLLB-3.3B model. (2) Flickr30k-200 -- 1,000 English captions from the test part of the Flickr30k dataset translated to 200 languages using NLLB-3.3B model.

\section{Training}

Since we already have a high-quality pre-trained text encoder, unlike other works \cite{carlsson2022cross, chen2022altclip}, we performed only fine-tuning on the collected dataset. This allowed us to minimize training costs per model and run many experiments within a limited budget.

On each epoch of training, we used only one randomly selected caption per image. It makes the training epochs shorter, and we don't need to run validation multiple times within an epoch to see the progress. Since the ratio between the training set size and the number of languages is ~447/1, we are confident that every language goes through the model in each epoch.

All experiments were performed on a single Nvidia H100 GPU with 80GB VRAM. Large memory capacity enabled using large batch sizes, which is crucial for contrastive loss \cite{radford2021learning}. We used Lion optimizer \cite{chen2023symbolic} as we found that it makes the model converge significantly faster and to better performance than weighted Adam \cite{loshchilov2017decoupled}. To save the GPU memory on the optimizer state, we used an 8-bit version of Lion \cite{dettmers2022optimizers}. This allowed us to have a 25\% larger batch size.

\section{Experiments}
\label{sec:experiments}

\subsection{Freezing different parts of NLLB-CLIP}
\label{sec:freeze}

\begin{figure}
  \includegraphics[width=12.5cm]{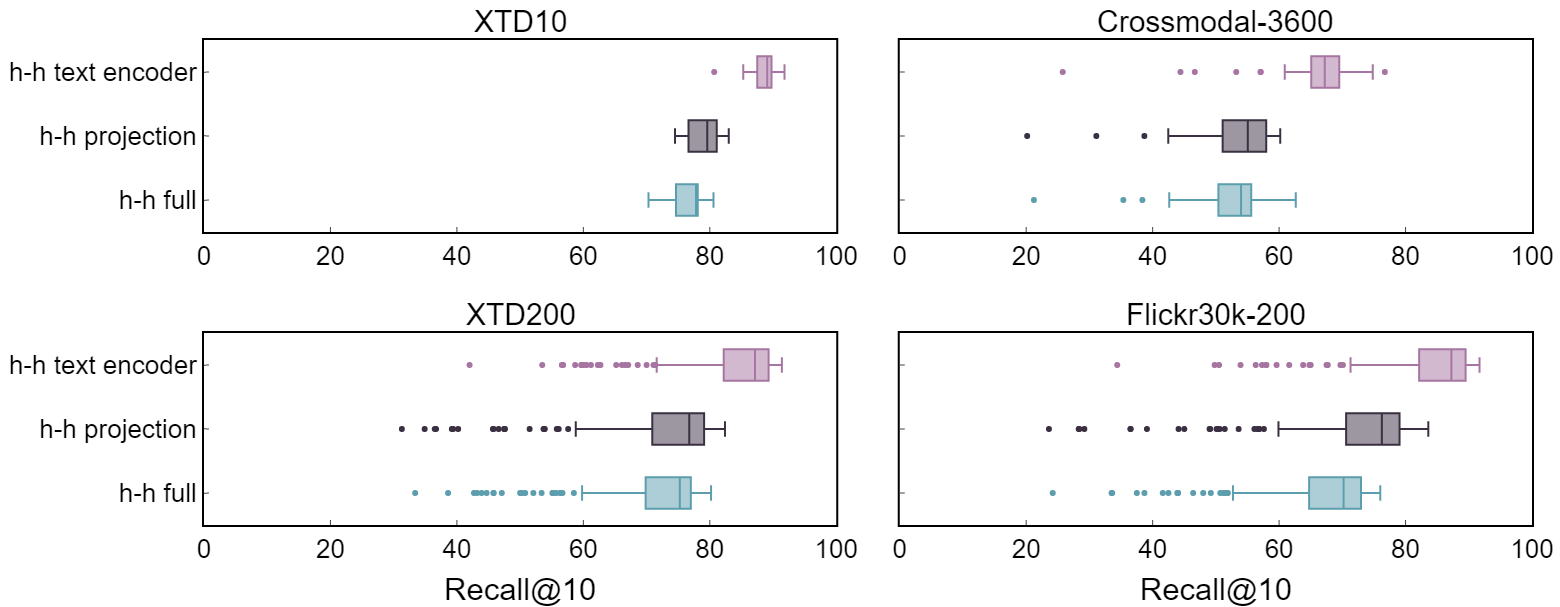}
  \centering
  \caption{ Performance on evaluation datasets for the same NLLB-CLIP model with different training regimes }
\label{fig:freeze}
\end{figure}

In this series of experiments, we explored the effect of freezing different parts of the model on its performance. We tried three training regimes: (1) freeze nothing (denoted as "full"); (2) freeze only image encoder and train text encoder and projection layers ("text encoder"); (3) freeze both image and text encoders and train only projection layers ("projection"). We tested all combinations of image and text encoder sizes and found that freezing only the image encoder produces significantly better results in all cases. Interestingly, training only projection layers gives better performance than full training. Exemplar results for the model with CLIP ViT huge and NLLB huge (h-h) are presented in Figure \ref{fig:freeze}. The smallest model variants (e.g., b-b and l-b) could not converge to get R@10 higher than 20\% when we performed training of the full model.

Our results are consistent with the results from the LiT paper \cite{zhai2022lit}, where the authors found that freezing the image encoder is the best regime for fine-tuning the model for the new tasks. Also, freezing the image encoder makes even more sense for our task, where for the same image, there are 201 different captions in the training dataset. The best training scenario in this case is to use high-quality image representations from the pre-trained image encoder to adjust the text encoder and align the text representations in the same latent space. To make this process faster, we train both visual and text projection layers along with the text encoder backbone.

\begin{figure}
  \includegraphics[width=11.5cm]{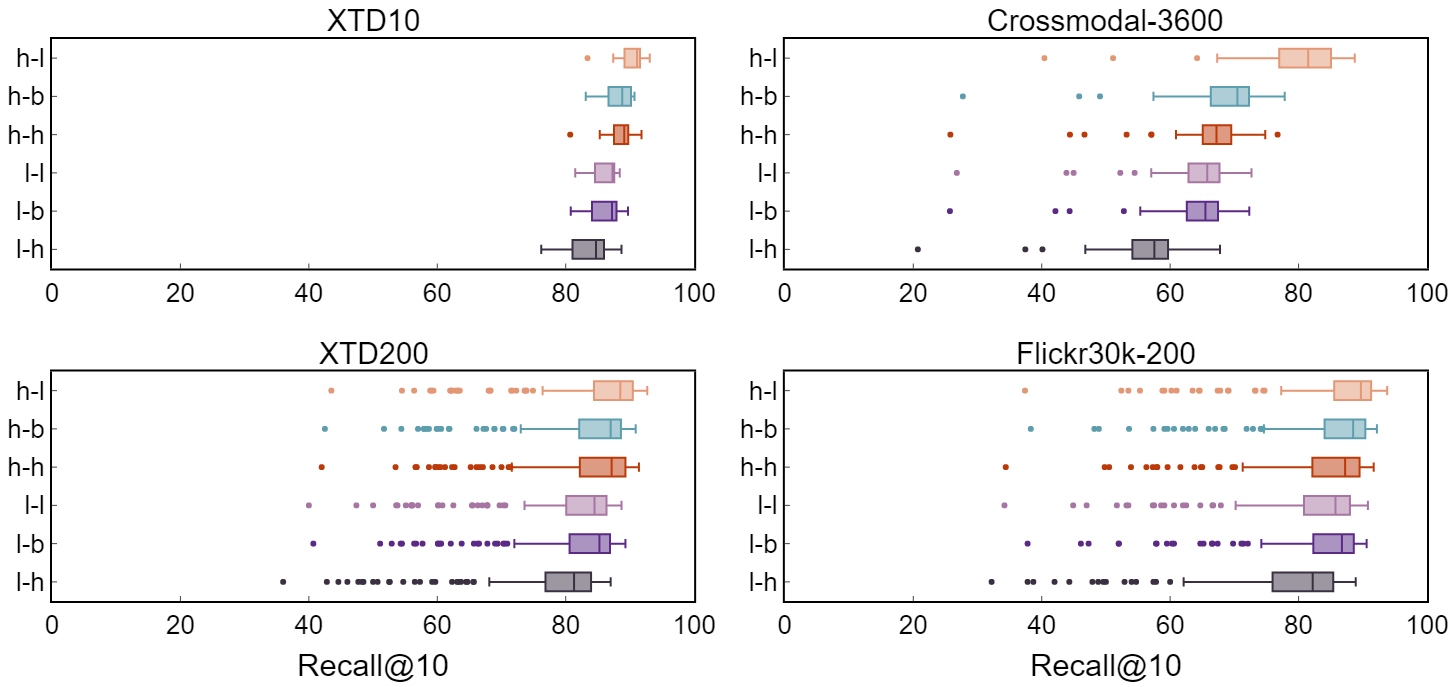}
  \centering
  \caption{ Performance on evaluation datasets for different combinations of image and text encoders }
\label{fig:size}
\end{figure}

\subsection{Encoder backbones size effect}
\label{sec:size}

In the next part of the experiments, we investigated the effect of backbone sizes on the model performance. For all models, we kept the image encoder frozen and trained only the text encoder with visual and text projection layers. We examined six models -- l-b, l-l, l-h, h-b, h-l, and h-h (refer to Section \ref{sec:nllb-clip} for abbreviation sources and respective backbone sizes). From the experiments (Figure \ref{fig:size}), we can see that for all datasets, the models with larger text encoders perform worse than the ones with smaller text encoders. For the large image encoder, the base text encoder performs better than the large one.

The results demonstrate that a larger (and higher quality) image encoder enables better results for any text encoder. Regarding the worse performance of larger text encoders, we attribute this phenomenon to a lack of data to fully align the largest text encoder with the image encoder in the latent space. The need for a sufficient amount of data for model training has been discussed in the literature \cite{soekhoe2016size, linjordet2019size, ji2023size}. We plan to collect more data and train the models to validate this hypothesis.

\section{Evaluation}
\label{sec:evaluation}

For comparison with existing works, we used two variants of NLLB-CLIP: (1) CLIP ViT base with NLLB base -- NLLB-CLIP base (501M parameters) and (2) CLIP ViT huge with NLLB large -- NLLB-CLIP large (1.4B parameters). Both models were trained with an image encoder frozen. We chose NLLB-CLIP large because it is the best-performing variant across all experiments. NLLB-CLIP base was chosen to evaluate the capabilities of the smallest model.

\begin{figure}
  \includegraphics[width=12cm]{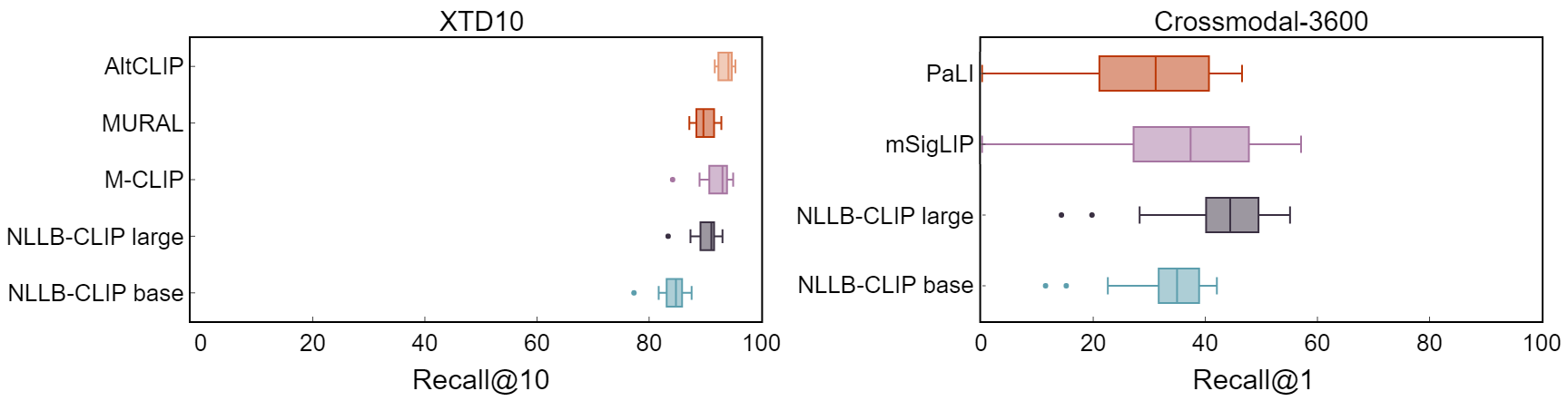}
  \centering
  \caption{ Evaluation results on multilingual datasets }
  \label{fig:evaluation}
\end{figure}

\subsection{XTD10}
\label{sec:xtd10}

For the XTD10 dataset, we used Recall@10 as a comparison metric, as it is used in other works. We compared NLLB-CLIP with state-of-the-art models -- Multilingual CLIP \cite{carlsson2022cross}, MURAL \cite{jain2021mural}, and AltCLIP \cite{chen2022altclip}. Figure \ref{fig:evaluation} (left) shows the results of the experiments. Although NLLB-CLIP did not outperform state-of-the-art models, the large model is not very far behind - 90.1\% vs. 93.7\% on average. It is worth mentioning that other models were trained on significantly larger datasets (up to 1,000x).

\subsection{Crossmodal-3600}
\label{sec:xm3600}

For the Crossmodal-3600 dataset, we used Recall@1 as a comparison metric, as it is used in other works. We compared NLLB-CLIP with state-of-the-art models -- mSigLIP \cite{zhai2023sigmoid} and PaLI \cite{chen2022pali}. Figure \ref{fig:evaluation} (right) shows the evaluation results. NLLB-CLIP large sets state-of-the-art results with 42.96\% R@1 on average across 36 languages. mSigLIP outperforms NLLB-CLIP in high-resource languages like English, Italian, or Spanish. The contrary is true for lower- and low-resource languages -- the advantage of NLLB-CLIP is higher the lower the resourcefulness of the target language. These results demonstrate the advantage of using the LAION-COCO-NLLB dataset, where all languages are represented equally, no matter their real-world resourcefulness.

\section{Conclusion}

In this paper, we demonstrate that by replacing the text encoder and fine-tuning on a small automatically created dataset, we can create a CLIP model capable of high-quality image retrieval in 201 languages of the Flores-200 dataset. NLLB-CLIP performs better than existing models on low-resource languages, primarily because the training dataset has the same number of captions for both low- and high-resource languages. NLLB-CLIP large sets new state-of-the-art results on the Crossmodal-3600 dataset that includes low- and high-resource languages.

\begin{ack}

We thank Lambda \footnote{\url{https://lambdalabs.com/}} for providing compute credits for running data collection and experiments.

We thank the ML Collective community for helpful discussions, ideas, and feedback on experiments.
\end{ack}

\bibliographystyle{abbrvnat}
\bibliography{bibliography}

\begin{thebibliography}{32}
\providecommand{\natexlab}[1]{#1}
\providecommand{\url}[1]{\texttt{#1}}
\expandafter\ifx\csname urlstyle\endcsname\relax
  \providecommand{\doi}[1]{doi: #1}\else
  \providecommand{\doi}{doi: \begingroup \urlstyle{rm}\Url}\fi

\bibitem[Aggarwal and Kale(2020)]{aggarwal2020zeroshot}
P.~Aggarwal and A.~Kale.
\newblock Towards zero-shot cross-lingual image retrieval, 2020.

\bibitem[Bianchi et~al.(2021)Bianchi, Attanasio, Pisoni, Terragni, Sarti, and
  Lakshmi]{bianchi2021contrastive}
F.~Bianchi, G.~Attanasio, R.~Pisoni, S.~Terragni, G.~Sarti, and S.~Lakshmi.
\newblock Contrastive language-image pre-training for the italian language.
\newblock \emph{arXiv preprint arXiv:2108.08688}, 2021.

\bibitem[Carlsson et~al.(2022)Carlsson, Eisen, Rekathati, and
  Sahlgren]{carlsson2022cross}
F.~Carlsson, P.~Eisen, F.~Rekathati, and M.~Sahlgren.
\newblock Cross-lingual and multilingual clip.
\newblock In \emph{Proceedings of the Thirteenth Language Resources and
  Evaluation Conference}, pages 6848--6854, 2022.

\bibitem[Chen et~al.(2022{\natexlab{a}})Chen, Wang, Changpinyo, Piergiovanni,
  Padlewski, Salz, Goodman, Grycner, Mustafa, Beyer, et~al.]{chen2022pali}
X.~Chen, X.~Wang, S.~Changpinyo, A.~Piergiovanni, P.~Padlewski, D.~Salz,
  S.~Goodman, A.~Grycner, B.~Mustafa, L.~Beyer, et~al.
\newblock Pali: A jointly-scaled multilingual language-image model.
\newblock \emph{arXiv preprint arXiv:2209.06794}, 2022{\natexlab{a}}.

\bibitem[Chen et~al.(2023)Chen, Liang, Huang, Real, Wang, Liu, Pham, Dong,
  Luong, Hsieh, et~al.]{chen2023symbolic}
X.~Chen, C.~Liang, D.~Huang, E.~Real, K.~Wang, Y.~Liu, H.~Pham, X.~Dong,
  T.~Luong, C.-J. Hsieh, et~al.
\newblock Symbolic discovery of optimization algorithms.
\newblock \emph{arXiv preprint arXiv:2302.06675}, 2023.

\bibitem[Chen et~al.(2022{\natexlab{b}})Chen, Liu, Zhang, Ye, Yang, and
  Wu]{chen2022altclip}
Z.~Chen, G.~Liu, B.-W. Zhang, F.~Ye, Q.~Yang, and L.~Wu.
\newblock Altclip: Altering the language encoder in clip for extended language
  capabilities.
\newblock \emph{arXiv preprint arXiv:2211.06679}, 2022{\natexlab{b}}.

\bibitem[Costa-juss{\`a} et~al.(2022)Costa-juss{\`a}, Cross, {\c{C}}elebi,
  Elbayad, Heafield, Heffernan, Kalbassi, Lam, Licht, Maillard,
  et~al.]{costa2022no}
M.~R. Costa-juss{\`a}, J.~Cross, O.~{\c{C}}elebi, M.~Elbayad, K.~Heafield,
  K.~Heffernan, E.~Kalbassi, J.~Lam, D.~Licht, J.~Maillard, et~al.
\newblock No language left behind: Scaling human-centered machine translation.
\newblock \emph{arXiv preprint arXiv:2207.04672}, 2022.

\bibitem[Dettmers et~al.(2022)Dettmers, Lewis, Shleifer, and
  Zettlemoyer]{dettmers2022optimizers}
T.~Dettmers, M.~Lewis, S.~Shleifer, and L.~Zettlemoyer.
\newblock 8-bit optimizers via block-wise quantization.
\newblock \emph{9th International Conference on Learning Representations,
  ICLR}, 2022.

\bibitem[Dosovitskiy et~al.(2020)Dosovitskiy, Beyer, Kolesnikov, Weissenborn,
  Zhai, Unterthiner, Dehghani, Minderer, Heigold, Gelly,
  et~al.]{dosovitskiy2020image}
A.~Dosovitskiy, L.~Beyer, A.~Kolesnikov, D.~Weissenborn, X.~Zhai,
  T.~Unterthiner, M.~Dehghani, M.~Minderer, G.~Heigold, S.~Gelly, et~al.
\newblock An image is worth 16x16 words: Transformers for image recognition at
  scale.
\newblock \emph{arXiv preprint arXiv:2010.11929}, 2020.

\bibitem[Jain et~al.(2021)Jain, Guo, Srinivasan, Chen, Kudugunta, Jia, Yang,
  and Baldridge]{jain2021mural}
A.~Jain, M.~Guo, K.~Srinivasan, T.~Chen, S.~Kudugunta, C.~Jia, Y.~Yang, and
  J.~Baldridge.
\newblock Mural: Multimodal, multitask representations across languages.
\newblock In \emph{Findings of the Association for computational Linguistics:
  EMNLP 2021}, pages 3449--3463, 2021.

\bibitem[Ji et~al.(2023)Ji, Deng, Gong, Peng, Niu, Zhang, Ma, and
  Li]{ji2023size}
Y.~Ji, Y.~Deng, Y.~Gong, Y.~Peng, Q.~Niu, L.~Zhang, B.~Ma, and X.~Li.
\newblock Exploring the impact of instruction data scaling on large language
  models: An empirical study on real-world use cases.
\newblock \emph{arXiv preprint arXiv:2303.14742}, 2023.

\bibitem[Joshi et~al.(2020)Joshi, Santy, Budhiraja, Bali, and
  Choudhury]{joshi2020state}
P.~Joshi, S.~Santy, A.~Budhiraja, K.~Bali, and M.~Choudhury.
\newblock The state and fate of linguistic diversity and inclusion in the nlp
  world.
\newblock \emph{arXiv preprint arXiv:2004.09095}, 2020.

\bibitem[Li et~al.(2022)Li, Shakhnarovich, and Yeh]{li2022adapting}
J.~Li, G.~Shakhnarovich, and R.~A. Yeh.
\newblock Adapting clip for phrase localization without further training.
\newblock \emph{arXiv preprint arXiv:2204.03647}, 2022.

\bibitem[Li et~al.(2019)Li, Xu, Wang, Lan, Jia, Yang, and Xu]{li2019coco}
X.~Li, C.~Xu, X.~Wang, W.~Lan, Z.~Jia, G.~Yang, and J.~Xu.
\newblock Coco-cn for cross-lingual image tagging, captioning, and retrieval.
\newblock \emph{IEEE Transactions on Multimedia}, 21\penalty0 (9):\penalty0
  2347--2360, 2019.

\bibitem[Lin et~al.(2014)Lin, Maire, Belongie, Hays, Perona, Ramanan,
  Doll{\'a}r, and Zitnick]{lin2014microsoft}
T.-Y. Lin, M.~Maire, S.~Belongie, J.~Hays, P.~Perona, D.~Ramanan,
  P.~Doll{\'a}r, and C.~L. Zitnick.
\newblock Microsoft coco: Common objects in context.
\newblock In \emph{Computer Vision--ECCV 2014: 13th European Conference,
  Zurich, Switzerland, September 6-12, 2014, Proceedings, Part V 13}, pages
  740--755. Springer, 2014.

\bibitem[Lin et~al.(2023)Lin, Chen, Wang, Wu, Li, Lin, Liu, and
  He]{lin2023clip}
Y.~Lin, M.~Chen, W.~Wang, B.~Wu, K.~Li, B.~Lin, H.~Liu, and X.~He.
\newblock Clip is also an efficient segmenter: A text-driven approach for
  weakly supervised semantic segmentation.
\newblock In \emph{Proceedings of the IEEE/CVF Conference on Computer Vision
  and Pattern Recognition}, pages 15305--15314, 2023.

\bibitem[Linjordet and Balog(2019)]{linjordet2019size}
T.~Linjordet and K.~Balog.
\newblock Impact of training dataset size on neural answer selection models.
\newblock In \emph{Advances in Information Retrieval: 41st European Conference
  on IR Research, ECIR 2019, Cologne, Germany, April 14--18, 2019, Proceedings,
  Part I 41}, pages 828--835. Springer, 2019.

\bibitem[Loshchilov and Hutter(2017)]{loshchilov2017decoupled}
I.~Loshchilov and F.~Hutter.
\newblock Decoupled weight decay regularization.
\newblock \emph{arXiv preprint arXiv:1711.05101}, 2017.

\bibitem[Nekoto et~al.(2020)Nekoto, Marivate, Matsila, Fasubaa, Kolawole,
  Fagbohungbe, Akinola, Muhammad, Kabongo, Osei,
  et~al.]{nekoto2020participatory}
W.~Nekoto, V.~Marivate, T.~Matsila, T.~Fasubaa, T.~Kolawole, T.~Fagbohungbe,
  S.~O. Akinola, S.~H. Muhammad, S.~Kabongo, S.~Osei, et~al.
\newblock Participatory research for low-resourced machine translation: A case
  study in african languages.
\newblock \emph{arXiv preprint arXiv:2010.02353}, 2020.

\bibitem[Radford et~al.(2021)Radford, Kim, Hallacy, Ramesh, Goh, Agarwal,
  Sastry, Askell, Mishkin, Clark, et~al.]{radford2021learning}
A.~Radford, J.~W. Kim, C.~Hallacy, A.~Ramesh, G.~Goh, S.~Agarwal, G.~Sastry,
  A.~Askell, P.~Mishkin, J.~Clark, et~al.
\newblock Learning transferable visual models from natural language
  supervision.
\newblock In \emph{International conference on machine learning}, pages
  8748--8763. PMLR, 2021.

\bibitem[Rajendran et~al.(2016)Rajendran, Khapra, Chandar, and
  Ravindran]{rajendran-etal-2016-bridge}
J.~Rajendran, M.~M. Khapra, S.~Chandar, and B.~Ravindran.
\newblock Bridge correlational neural networks for multilingual multimodal
  representation learning.
\newblock In \emph{Proceedings of the 2016 Conference of the North {A}merican
  Chapter of the Association for Computational Linguistics: Human Language
  Technologies}, pages 171--181, San Diego, California, June 2016. Association
  for Computational Linguistics.
\newblock \doi{10.18653/v1/N16-1021}.
\newblock URL \url{https://aclanthology.org/N16-1021}.

\bibitem[Soekhoe et~al.(2016)Soekhoe, Van Der~Putten, and
  Plaat]{soekhoe2016size}
D.~Soekhoe, P.~Van Der~Putten, and A.~Plaat.
\newblock On the impact of data set size in transfer learning using deep neural
  networks.
\newblock In \emph{Advances in Intelligent Data Analysis XV: 15th International
  Symposium, IDA 2016, Stockholm, Sweden, October 13-15, 2016, Proceedings 15},
  pages 50--60. Springer, 2016.

\bibitem[Thapliyal et~al.(2022)Thapliyal, Pont-Tuset, Chen, and
  Soricut]{ThapliyalCrossmodal2022}
A.~Thapliyal, J.~Pont-Tuset, X.~Chen, and R.~Soricut.
\newblock {Crossmodal-3600: A Massively Multilingual Multimodal Evaluation
  Dataset}.
\newblock In \emph{EMNLP}, 2022.

\bibitem[Thengane et~al.(2022)Thengane, Khan, Hayat, and
  Khan]{thengane2022clip}
V.~Thengane, S.~Khan, M.~Hayat, and F.~Khan.
\newblock Clip model is an efficient continual learner.
\newblock \emph{arXiv preprint arXiv:2210.03114}, 2022.

\bibitem[Vaswani et~al.(2017)Vaswani, Shazeer, Parmar, Uszkoreit, Jones, Gomez,
  Kaiser, and Polosukhin]{vaswani2017attention}
A.~Vaswani, N.~Shazeer, N.~Parmar, J.~Uszkoreit, L.~Jones, A.~N. Gomez,
  {\L}.~Kaiser, and I.~Polosukhin.
\newblock Attention is all you need.
\newblock \emph{Advances in neural information processing systems}, 30, 2017.

\bibitem[Visheratin(2023)]{laion-coco-nllb}
A.~Visheratin.
\newblock Laion-coco-nllb dataset, 2023.
\newblock URL \url{https://huggingface.co/datasets/visheratin/laion-coco-nllb}.

\bibitem[Xie et~al.(2022)Xie, Cai, Song, Li, Kong, Wu, Morimitsu, Yao, Wang,
  Zhang, Leng, Ji, and Deng]{xie2022zero}
C.~Xie, H.~Cai, J.~Song, J.~Li, F.~Kong, X.~Wu, H.~Morimitsu, L.~Yao, D.~Wang,
  X.~Zhang, D.~Leng, X.~Ji, and Y.~Deng.
\newblock Zero and r2d2: A large-scale chinese cross-modal benchmark and a
  vision-language framework, 2022.

\bibitem[Yang et~al.(2022)Yang, Pan, Lin, Men, Zhang, Zhou, and
  Zhou]{yang2022chinese}
A.~Yang, J.~Pan, J.~Lin, R.~Men, Y.~Zhang, J.~Zhou, and C.~Zhou.
\newblock Chinese clip: Contrastive vision-language pretraining in chinese.
\newblock \emph{arXiv preprint arXiv:2211.01335}, 2022.

\bibitem[Yoshikawa et~al.(2017)Yoshikawa, Shigeto, and
  Takeuchi]{yoshikawa-etal-2017-stair}
Y.~Yoshikawa, Y.~Shigeto, and A.~Takeuchi.
\newblock {STAIR} captions: Constructing a large-scale {J}apanese image caption
  dataset.
\newblock In \emph{Proceedings of the 55th Annual Meeting of the Association
  for Computational Linguistics (Volume 2: Short Papers)}, pages 417--421,
  Vancouver, Canada, July 2017. Association for Computational Linguistics.
\newblock \doi{10.18653/v1/P17-2066}.
\newblock URL \url{https://aclanthology.org/P17-2066}.

\bibitem[Young et~al.(2014)Young, Lai, Hodosh, and Hockenmaier]{young2014image}
P.~Young, A.~Lai, M.~Hodosh, and J.~Hockenmaier.
\newblock From image descriptions to visual denotations: New similarity metrics
  for semantic inference over event descriptions.
\newblock \emph{Transactions of the Association for Computational Linguistics},
  2:\penalty0 67--78, 2014.

\bibitem[Zhai et~al.(2022)Zhai, Wang, Mustafa, Steiner, Keysers, Kolesnikov,
  and Beyer]{zhai2022lit}
X.~Zhai, X.~Wang, B.~Mustafa, A.~Steiner, D.~Keysers, A.~Kolesnikov, and
  L.~Beyer.
\newblock Lit: Zero-shot transfer with locked-image text tuning.
\newblock In \emph{Proceedings of the IEEE/CVF Conference on Computer Vision
  and Pattern Recognition}, pages 18123--18133, 2022.

\bibitem[Zhai et~al.(2023)Zhai, Mustafa, Kolesnikov, and
  Beyer]{zhai2023sigmoid}
X.~Zhai, B.~Mustafa, A.~Kolesnikov, and L.~Beyer.
\newblock Sigmoid loss for language image pre-training.
\newblock \emph{arXiv preprint arXiv:2303.15343}, 2023.

\end{thebibliography}

\clearpage

\appendix

\section{Additional experiments}

We performed additional experiments to evaluate the performance of NLLB-CLIP models on single-language datasets. We used the following datasets:

\begin{enumerate}
    \item COCO-CN \cite{li2019coco} (test) - 1,000 image-text pairs from the MS COCO 2014 dataset manually translated into Chinese.
    \item MSCOCO-IT \cite{bianchi2021contrastive} (validation) - 8,485 captions for 1,697 images from the MS COCO 2014 dataset translated into Italian.
    \item Flickr30k \cite{young2014image} (test) - 1,000 image-text pairs from Flickr30k dataset in English.
    \item Flickr30k-CN \cite{xie2022zero} - test part of the Flickr30k dataset (1,000 image-text pairs) manually translated to Chinese.
\end{enumerate}

Table \ref{tab:coco_flickr30k} shows the results of the experiments on COCO-CN, Flickr30k-CN, and Flickr30k-EN datasets. Since this version of AltCLIP was trained in two stages and targeted only English and Chinese, it yields noticeably higher results in all metrics. Both bilingual and multilingual results of AltCLIP show that the two-stage training is highly beneficial for aligning image and text representations. We plan to explore this approach in the future.

\begin{table}
  \caption{COCO CN and Flickr30k datasets evaluation results}
  \label{tab:coco_flickr30k}
  \centering
  \begin{tabular}{lccccccccc}
    \toprule
     & \multicolumn{3}{c}{NLLB-CLIP base} & \multicolumn{3}{c}{NLLB-CLIP large} & \multicolumn{3}{c}{AltCLIP} \\
    \cmidrule(lr){2-4}
    \cmidrule(lr){5-7}
    \cmidrule(lr){8-10}
    Dataset & R@1 & R@5 & R@10 & R@1 & R@5 & R@10 & R@1 & R@5 & R@10 \\
    \midrule
    COCO CN       & 45.21 & 76.77 & 87.51 & 53.05 & 82.67 & 92.64 & \textbf{63.9} & \textbf{87.2} & \textbf{93.9} \\
    Flickr30k CN  & 38.36 & 66.72 & 76.0 & 50.92 & 77.18 & 85.28 & \textbf{69.8} & \textbf{89.9} & \textbf{94.7} \\
    Flickr30k     & 54.6 & 81.2 & 86.9 & 66.9 & 89.4  & 93.5  & \textbf{72.5} & \textbf{91.6} & \textbf{95.4} \\
    \bottomrule
  \end{tabular}
\end{table}

Table \ref{tab:coco_it} shows the evaluation results on the COCO-IT dataset. To perform a fair comparison, we used the Jupyter notebooks the authors provided and changed only the parts related to loading and running the model. We used Mean Reciprocal Rank (MRR) as a comparison metric because it was used in the original paper \cite{bianchi2021contrastive}. NLLB-CLIP large outperforms the original results and sets state-of-the-art on the COCO-IT dataset.

\begin{table}
  \caption{COCO Italian dataset evaluation results}
  \label{tab:coco_it}
  \centering
  \begin{tabular}{lccccccccc}
    \toprule
     & \multicolumn{3}{c}{NLLB-CLIP base} & \multicolumn{3}{c}{NLLB-CLIP large} & \multicolumn{3}{c}{Italian CLIP} \\
    \cmidrule(lr){2-4}
    \cmidrule(lr){5-7}
    \cmidrule(lr){8-10}
    Dataset & @1 & @5 & @10 & @1 & @5 & @10 & @1 & @5 & @10 \\
    \midrule
    COCO IT & 35.5  & 47.1  & 48.7  & \textbf{45.4}  & \textbf{56.7}  & \textbf{58.1}   & 37.97 & 50.39 & 52.04 \\
    \bottomrule
  \end{tabular}
\end{table}

\section{XTD10 results}

Table \ref{tab:xtd10} shows the detailed results of the experiments presented in Section \ref{sec:xtd10}.

\begin{table}
  \caption{XTD10 dataset evaluation results}
  \label{tab:xtd10}
  \centering
  \begin{tabular}{lccccccccc}
    \toprule
     & \multicolumn{3}{c}{NLLB-CLIP base} & \multicolumn{3}{c}{NLLB-CLIP large} & \multicolumn{1}{c}{M-CLIP} & \multicolumn{1}{c}{MURAL} & \multicolumn{1}{c}{AltCLIP} \\
    \cmidrule(lr){2-4}
    \cmidrule(lr){5-7}
    \cmidrule(lr){8-8}
    \cmidrule(lr){9-9}
    \cmidrule(lr){10-10}
    Language & R@1 & R@5 & R@10 & R@1 & R@5 & R@10 & R@10 & R@10 & R@10 \\
    \midrule
    en & 47.2 & 76.9 & 87.6 & 59.3 & 83.6 & 93.1 & 95.0 & - & \textbf{95.4} \\
    fr & 45.6 & 74.8 & 84.8 & 56.7 & 82.3 & 91.1 & \textbf{93.1} & - & 92.9 \\
    zh & 42.4 & 73.9 & 85.5 & 52.7 & 79.9 & 90.5 & 94.0 & 89.7 & \textbf{95.1} \\
    de & 46.4 & 74.4 & 84.8 & 56.2 & 82.2 & 90.6 & \textbf{93.1} & - & - \\
    es & 44.9 & 75.4 & 85.9 & 54.9 & 81.8 & 91.7 & 93.6 & 92.9 & \textbf{94.1} \\
    it & 44.5 & 75.1 & 85.9 & 54.8 & 81.7 & 91.2 & 93.1 & 91.8 & \textbf{94.2} \\
    tr & 43.5 & 72.8 & 84.2 & 55.8 & 81.6 & 91.2 & \textbf{93.0}  & 89.5 & - \\
    ru & 41.5 & 70.4 & 81.7 & 51.7 & 77.3 & 87.4 & 90.0   & 87.2 & \textbf{91.8} \\
    ko & 40.1 & 70.5 & 82.8 & 50.2 & 77.2 & 88.7 & 89.0  & 88.1 & \textbf{94.4} \\
    pl & 45.3 & 75.4 & 86.2 & 56.2 & 82.9 & 91.9 & \textbf{94.4} & 91.0  & - \\
    jp & 38.1 & 67.3 & 77.3 & 47.6 & 74.3 & 83.4 & 84.2 & -     & \textbf{91.7} \\
    \bottomrule
  \end{tabular}
\end{table}

\section{Crossmodal-3600 results}

Table \ref{tab:xm3600} shows the detailed results of the experiments presented in Section \ref{sec:xm3600}. The crossmodal-3600 dataset contains multiple captions per image. We used the first caption for the tests.

\begin{table}
  \caption{Crossmodal-3600 dataset evaluation results}
  \label{tab:xm3600}
  \centering
  \begin{tabular}{lcccccccc}
    \toprule
     & \multicolumn{3}{c}{NLLB-CLIP base} & \multicolumn{3}{c}{NLLB-CLIP large} & \multicolumn{1}{c}{mSigLIP} & \multicolumn{1}{c}{PaLI} \\
    \cmidrule(lr){2-4}
    \cmidrule(lr){5-7}
    \cmidrule(lr){8-8}
    \cmidrule(lr){9-9}
    Language & R@1 & R@5 & R@10 & R@1 & R@5 & R@10 & R@1 & R@1 \\
    \midrule
    ar  & 34.61 & 63.44 & 74.5  & \textbf{42.53} & 70.72 & 81.14 & 37.37 & 32.6 \\
    bn  & 32.36 & 60.0    & 70.83 & \textbf{40.25} & 66.36 & 76.78 & 6.25  & 3.31 \\
    cs  & 29.47 & 57.39 & 70.44 & 37.22 & 65.81 & 77.19 & \textbf{41.64} & 35.24 \\
    da  & 41.72 & 70.14 & 80.17 & \textbf{54.06} & 80.5  & 87.89 & 47.01 & 38.48 \\
    de  & 41.78 & 71.17 & 80.61 & \textbf{55.11} & 80.22 & 88.14 & 54.83 & 46.5  \\
    el & 31.0 & 56.94 & 68.86 & \textbf{40.03} & 67.67 & 78.03 & 22.78 & 20.92 \\
    en & 30.53 & 58.81 & 69.47 & 35.64 & 64.31 & 74.94 & \textbf{46.21} & 40.63 \\
    es & 36.03 & 64.86 & 75.92 & 45.25 & 73.0 & 82.64 & \textbf{55.04} & 46.55 \\
    fa & 34.28 & 61.14 & 71.5 & \textbf{41.44} & 69.97 & 79.17 & 40.15 & 35.58 \\
    fi & 37.33 & 65.11 & 75.28 & \textbf{48.06} & 75.33 & 84.17 & 37.14 & 21.8 \\
    fil & 25.06 & 50.83 & 62.58 & \textbf{32.47} & 58.33 & 69.31 & 12.93 & 10.04 \\
    fr & 40.58 & 69.94 & 80.03 & 52.11 & 78.92 & 86.58 & \textbf{57.08} & 43.47 \\
    he & 40.17 & 68.67 & 78.14 & \textbf{51.94} & 77.42 & 85.72 & 31.76 & 29.39 \\
    hi & 22.61 & 47.03 & 58.78 & \textbf{28.28} & 53.44 & 64.19 & 8.55 & 7.42 \\
    hr & 42.06 & 70.69 & 80.94 & \textbf{54.42} & 81.25 & 88.75 & 37.09 & 29.55 \\
    hu & 38.61 & 65.67 & 75.92 & \textbf{49.75} & 76.69 & 85.5 & 40.2 & 35.49 \\
    id & 41.42 & 68.53 & 79.42 & \textbf{50.44} & 77.22 & 85.06 & 49.42 & 36.75 \\
    it & 38.44 & 66.97 & 77.56 & 49.53 & 75.67 & 84.42 & \textbf{56.39} & 44.76 \\
    ja & 38.92 & 67.19 & 77.33 & \textbf{49.44} & 76.14 & 83.89 & 31.32 & 27.24 \\
    ko & 32.83 & 59.25 & 69.47 & \textbf{41.69} & 67.53 & 76.72 & 34.72 & 25.95 \\
    mi & 15.19 & 35.03 & 45.06 & \textbf{19.78} & 40.92 & 51.11 & 0.22 & 0.22 \\
    nl & 35.58 & 62.28 & 73.44 & 43.81 & 72.14 & 81.81 & \textbf{49.55} & 43.79 \\
    no & 34.75 & 62.06 & 72.69 & 43.06 & 71.17 & 80.17 & \textbf{46.21} & 37.35 \\
    pl & 34.89 & 62.94 & 73.47 & 45.28 & 72.5 & 81.19 & \textbf{47.36} & 43.72 \\
    pt & 35.03 & 62.72 & 74.25 & 45.06 & 73.47 & 82.36 & \textbf{52.34} & 42.73 \\
    quz & 11.5 & 25.17 & 33.89 & \textbf{14.33} & 31.22 & 40.42 & 2.57 & 1.9 \\
    ro & 41.14 & 70.0 & 80.44 & \textbf{53.36} & 80.56 & 88.42 & 35.6 & 28.82 \\
    ru & 39.78 & 68.5 & 78.78 & 48.58 & 77.0 & 85.58 & \textbf{49.89} & 41.11 \\
    sv & 33.44 & 60.97 & 72.39 & 41.58 & 69.28 & 78.92 & \textbf{48.18} & 40.66 \\
    sw & 27.06 & 54.0 & 66.0 & \textbf{34.5} & 62.61 & 72.83 & 7.17 & 3.41 \\
    te & 24.25 & 50.44 & 62.67 & \textbf{29.53} & 56.47 & 67.33 & 0.29 & 1.42 \\
    th & 33.53 & 62.11 & 73.06 & \textbf{42.25} & 70.03 & 79.78 & 23.08 & 16.06 \\
    tr & 36.92 & 65.72 & 76.19 & \textbf{47.78} & 74.19 & 83.14 & 37.38 & 31.47 \\
    uk & 38.14 & 66.14 & 76.69 & \textbf{48.11} & 74.58 & 83.31 & 33.21 & 30.81 \\
    vi & 38.86 & 67.31 & 77.89 & \textbf{48.33} & 76.89 & 85.06 & 41.92 & 21.28 \\
    zh & 33.89 & 60.69 & 71.92 & \textbf{41.69} & 67.89 & 77.81 & 32.45 & 28.24 \\
    \midrule
    Average & 33.99 & 61.11 & 71.85 & \textbf{42.96} & 69.65 & 78.87 & 34.87 & 28.46 \\
    \bottomrule
  \end{tabular}
\end{table}

\section{Flores-200 datasets results}

Tables \ref{tab:xtd_200} and \ref{tab:flickr30k_200} present the results of NLLB-CLIP base and large models on XTD200 and Flickr30k-200 datasets. Datasets description is provided in Section \ref{sec:eval-datasets}.

\begin{longtable}{lcccccc}
    \caption{XTD200 evaluation dataset results} 
    \label{tab:xtd_200} \\
    \toprule
     & \multicolumn{3}{c}{NLLB-CLIP base} & \multicolumn{3}{c}{NLLB-CLIP large} \\
    \cmidrule(lr){2-4}
    \cmidrule(lr){5-7}
    Language & R@1 & R@5 & R@10 & R@1 & R@5 & R@10 \\
    \midrule
    \endfirsthead 

    \multicolumn{7}{c}{{\bfseries \tablename\ \thetable{} -- continued from the previous page}} \\
    \midrule
    & \multicolumn{3}{c}{NLLB-CLIP base} & \multicolumn{3}{c}{NLLB-CLIP large} \\
    \cmidrule(lr){2-4}
    \cmidrule(lr){5-7}
    Language & R@1 & R@5 & R@10 & R@1 & R@5 & R@10 \\
    \midrule
    \endhead 

    \midrule \multicolumn{7}{r}{{Continued on the next page}} \\
    \endfoot 

    \bottomrule
    \endlastfoot 

    ace\_Arab & 20.1 & 42.7 & 54.9 & 27.4 & 50.7 & 62.0 \\
    ace\_Latn & 40.3 & 70.2 & 82.3 & 50.6 & 76.7 & 86.8 \\
    acm\_Arab & 45.3 & 72.9 & 84.6 & 55.4 & 82.2 & 90.6 \\
    acq\_Arab & 45.5 & 73.4 & 84.0 & 55.5 & 81.6 & 90.3 \\
    aeb\_Arab & 33.8 & 64.1 & 75.2 & 45.9 & 68.8 & 79.2 \\
    afr\_Latn & 46.7 & 75.6 & 86.5 & 56.6 & 81.3 & 91.2 \\
    ajp\_Arab & 43.1 & 72.7 & 83.0 & 54.5 & 80.4 & 89.2 \\
    aka\_Latn & 33.1 & 60.1 & 74.2 & 42.7 & 69.6 & 81.4 \\
    als\_Latn & 44.3 & 74.8 & 85.2 & 55.1 & 81.0 & 91.0 \\
    amh\_Ethi & 42.3 & 72.1 & 84.0 & 51.8 & 78.1 & 88.7 \\
    apc\_Arab & 42.6 & 72.1 & 83.4 & 54.7 & 79.7 & 88.5 \\
    arb\_Arab & 45.1 & 74.7 & 85.0 & 57.3 & 83.3 & 90.7 \\
    ars\_Arab & 44.8 & 73.9 & 84.7 & 55.0 & 81.5 & 90.4 \\
    ary\_Arab & 40.8 & 69.1 & 80.0 & 49.7 & 76.8 & 85.6 \\
    arz\_Arab & 43.7 & 72.1 & 83.6 & 56.7 & 81.1 & 89.1 \\
    asm\_Beng & 42.2 & 71.3 & 82.6 & 51.7 & 79.1 & 89.1 \\
    ast\_Latn & 43.4 & 72.8 & 83.5 & 56.4 & 79.2 & 88.6 \\
    awa\_Deva & 39.0 & 67.1 & 79.2 & 48.5 & 73.6 & 83.2 \\
    ayr\_Latn & 35.4 & 63.4 & 75.9 & 43.4 & 72.2 & 82.4 \\
    azb\_Arab & 38.2 & 68.2 & 80.1 & 48.5 & 74.1 & 85.8 \\
    azj\_Latn & 43.2 & 75.8 & 85.6 & 55.1 & 80.6 & 89.7 \\
    bak\_Cyrl & 39.6 & 73.3 & 83.8 & 51.9 & 78.2 & 88.0 \\
    bam\_Latn & 32.1 & 60.1 & 71.7 & 39.4 & 66.5 & 77.6 \\
    ban\_Latn & 38.9 & 68.5 & 82.2 & 48.3 & 74.7 & 85.4 \\
    bel\_Cyrl & 44.0 & 74.6 & 85.2 & 55.4 & 82.7 & 91.1 \\
    bem\_Latn & 31.5 & 61.6 & 73.1 & 37.0 & 67.8 & 78.5 \\
    ben\_Beng & 43.4 & 74.6 & 84.9 & 54.8 & 80.9 & 90.5 \\
    bho\_Deva & 43.3 & 74.7 & 85.5 & 54.2 & 80.6 & 91.2 \\
    bjn\_Arab & 16.3 & 36.5 & 49.7 & 22.3 & 46.0 & 56.4 \\
    bjn\_Latn & 43.1 & 72.9 & 83.6 & 53.2 & 80.2 & 89.3 \\
    bod\_Tibt & 30.2 & 58.7 & 70.1 & 36.8 & 63.6 & 76.4 \\
    bos\_Latn & 44.2 & 76.1 & 86.1 & 56.3 & 81.5 & 91.5 \\
    bug\_Latn & 36.3 & 66.5 & 79.3 & 45.4 & 73.4 & 84.5 \\
    bul\_Cyrl & 44.1 & 75.8 & 85.9 & 56.3 & 81.5 & 91.1 \\
    cat\_Latn & 44.2 & 75.8 & 85.5 & 55.7 & 81.7 & 90.8 \\
    ceb\_Latn & 44.5 & 74.5 & 85.5 & 56.2 & 80.5 & 90.1 \\
    ces\_Latn & 43.9 & 76.5 & 86.4 & 57.2 & 81.7 & 90.3 \\
    cjk\_Latn & 22.4 & 44.7 & 58.0 & 28.4 & 52.0 & 63.5 \\
    ckb\_Arab & 42.0 & 71.4 & 83.1 & 52.2 & 79.3 & 88.4 \\
    crh\_Latn & 43.0 & 74.0 & 85.2 & 54.6 & 79.0 & 89.9 \\
    cym\_Latn & 38.6 & 68.3 & 80.6 & 49.3 & 76.5 & 87.0 \\
    dan\_Latn & 47.5 & 76.4 & 87.0 & 57.8 & 83.6 & 92.1 \\
    deu\_Latn & 46.4 & 77.2 & 86.4 & 58.3 & 83.1 & 91.9 \\
    dik\_Latn & 21.9 & 46.2 & 59.1 & 28.1 & 51.9 & 62.8 \\
    dyu\_Latn & 23.4 & 49.0 & 63.8 & 29.0 & 57.5 & 68.3 \\
    dzo\_Tibt & 27.7 & 56.7 & 68.8 & 35.7 & 62.4 & 73.9 \\
    ell\_Grek & 46.2 & 75.5 & 86.9 & 56.7 & 83.1 & 92.2 \\
    epo\_Latn & 45.5 & 74.8 & 86.4 & 55.7 & 81.7 & 90.9 \\
    est\_Latn & 43.2 & 71.5 & 82.3 & 52.9 & 78.8 & 88.5 \\
    eus\_Latn & 41.6 & 72.6 & 83.1 & 51.4 & 78.6 & 87.9 \\
    ewe\_Latn & 34.8 & 62.8 & 75.9 & 42.4 & 71.3 & 83.2 \\
    fao\_Latn & 42.9 & 71.4 & 83.7 & 53.8 & 79.3 & 89.4 \\
    fij\_Latn & 40.7 & 70.3 & 81.4 & 50.0 & 77.3 & 87.7 \\
    fin\_Latn & 43.3 & 71.5 & 82.1 & 52.3 & 78.3 & 88.8 \\
    fon\_Latn & 24.9 & 52.3 & 63.7 & 29.1 & 56.4 & 68.0 \\
    fra\_Latn & 46.0 & 77.5 & 87.3 & 58.0 & 82.2 & 91.7 \\
    fur\_Latn & 41.6 & 70.9 & 82.5 & 51.3 & 77.4 & 87.9 \\
    fuv\_Latn & 27.6 & 54.5 & 66.7 & 35.0 & 61.0 & 71.7 \\
    gaz\_Latn & 36.5 & 64.5 & 77.6 & 43.6 & 71.8 & 82.0 \\
    gla\_Latn & 40.4 & 71.0 & 81.7 & 50.4 & 78.8 & 87.4 \\
    gle\_Latn & 41.1 & 69.6 & 80.9 & 51.5 & 77.8 & 88.1 \\
    glg\_Latn & 45.3 & 76.4 & 86.6 & 57.9 & 81.9 & 92.2 \\
    grn\_Latn & 40.3 & 68.2 & 79.9 & 46.9 & 75.7 & 86.2 \\
    guj\_Gujr & 45.2 & 75.1 & 85.2 & 55.6 & 81.6 & 91.1 \\
    hat\_Latn & 43.3 & 74.6 & 85.1 & 54.9 & 80.4 & 89.8 \\
    hau\_Latn & 42.8 & 68.7 & 81.3 & 53.5 & 78.0 & 88.2 \\
    heb\_Hebr & 42.8 & 72.5 & 83.6 & 56.1 & 81.8 & 90.8 \\
    hin\_Deva & 45.1 & 76.5 & 86.2 & 57.1 & 82.9 & 91.6 \\
    hne\_Deva & 43.5 & 75.3 & 85.4 & 54.3 & 80.3 & 90.1 \\
    hrv\_Latn & 45.4 & 76.1 & 85.8 & 57.0 & 81.6 & 91.5 \\
    hun\_Latn & 44.2 & 76.0 & 85.8 & 54.8 & 81.9 & 90.9 \\
    hye\_Armn & 42.0 & 72.6 & 83.7 & 52.9 & 78.9 & 88.9 \\
    ibo\_Latn & 41.1 & 70.0 & 83.0 & 50.3 & 76.8 & 86.4 \\
    ilo\_Latn & 43.0 & 74.0 & 83.8 & 55.6 & 80.0 & 89.9 \\
    ind\_Latn & 47.3 & 75.4 & 84.8 & 56.0 & 82.0 & 91.1 \\
    isl\_Latn & 41.1 & 71.4 & 83.0 & 52.0 & 79.3 & 89.1 \\
    ita\_Latn & 45.3 & 76.6 & 87.5 & 56.8 & 82.7 & 92.6 \\
    jav\_Latn & 45.0 & 74.1 & 86.0 & 55.3 & 81.0 & 90.1 \\
    jpn\_Jpan & 44.3 & 73.7 & 84.9 & 53.9 & 81.0 & 90.5 \\
    kab\_Latn & 32.4 & 60.6 & 72.3 & 38.8 & 68.1 & 78.3 \\
    kac\_Latn & 30.4 & 57.3 & 69.3 & 37.5 & 63.1 & 74.9 \\
    kam\_Latn & 11.3 & 30.8 & 40.7 & 14.5 & 32.6 & 43.5 \\
    kan\_Knda & 44.0 & 74.9 & 85.3 & 57.2 & 81.9 & 91.4 \\
    kas\_Arab & 34.2 & 61.8 & 73.9 & 43.8 & 69.9 & 80.9 \\
    kas\_Deva & 39.3 & 70.4 & 79.6 & 48.5 & 74.7 & 85.8 \\
    kat\_Geor & 39.6 & 70.6 & 81.4 & 50.2 & 76.8 & 86.6 \\
    kaz\_Cyrl & 42.7 & 72.9 & 84.1 & 51.8 & 78.9 & 88.4 \\
    kbp\_Latn & 17.5 & 37.5 & 48.5 & 21.0 & 43.9 & 54.5 \\
    kea\_Latn & 40.4 & 68.8 & 80.7 & 49.7 & 74.5 & 85.7 \\
    khk\_Cyrl & 34.8 & 62.0 & 74.8 & 42.7 & 69.2 & 78.8 \\
    khm\_Khmr & 36.2 & 65.0 & 75.7 & 44.1 & 71.1 & 83.0 \\
    kik\_Latn & 31.1 & 53.4 & 65.7 & 37.1 & 61.5 & 73.6 \\
    kin\_Latn & 38.0 & 67.0 & 79.8 & 46.3 & 73.5 & 84.3 \\
    kir\_Cyrl & 39.4 & 72.8 & 84.6 & 51.7 & 77.6 & 88.0 \\
    kmb\_Latn & 19.9 & 41.6 & 54.0 & 22.4 & 46.2 & 58.9 \\
    kmr\_Latn & 40.4 & 70.1 & 81.5 & 51.6 & 78.4 & 88.8 \\
    knc\_Arab & 35.8 & 64.9 & 76.0 & 47.4 & 72.4 & 82.3 \\
    knc\_Latn & 24.9 & 49.5 & 64.4 & 29.6 & 54.3 & 68.2 \\
    kon\_Latn & 34.9 & 63.8 & 75.4 & 42.6 & 70.2 & 81.7 \\
    kor\_Hang & 42.5 & 72.6 & 83.5 & 53.6 & 80.3 & 90.0 \\
    lao\_Laoo & 41.9 & 71.6 & 84.4 & 50.5 & 76.9 & 86.6 \\
    lij\_Latn & 38.4 & 68.7 & 80.0 & 48.4 & 75.0 & 85.7 \\
    lim\_Latn & 45.3 & 75.0 & 85.5 & 54.5 & 81.3 & 91.8 \\
    lin\_Latn & 36.6 & 68.5 & 80.9 & 48.7 & 77.8 & 87.8 \\
    lit\_Latn & 41.5 & 72.1 & 82.8 & 51.7 & 78.6 & 89.1 \\
    lmo\_Latn & 43.8 & 73.3 & 85.2 & 55.1 & 81.3 & 91.5 \\
    ltg\_Latn & 35.1 & 63.9 & 75.3 & 43.1 & 70.0 & 79.5 \\
    ltz\_Latn & 45.7 & 75.7 & 85.5 & 56.5 & 81.2 & 90.9 \\
    lua\_Latn & 31.2 & 59.1 & 72.7 & 37.3 & 68.1 & 79.8 \\
    lug\_Latn & 34.8 & 63.6 & 76.0 & 41.3 & 70.6 & 82.1 \\
    luo\_Latn & 33.0 & 61.0 & 73.5 & 40.7 & 69.2 & 80.9 \\
    lus\_Latn & 33.6 & 60.0 & 74.0 & 41.4 & 66.9 & 79.3 \\
    lvs\_Latn & 40.2 & 69.5 & 81.4 & 50.2 & 76.1 & 86.8 \\
    mag\_Deva & 45.1 & 76.1 & 86.6 & 56.2 & 82.3 & 91.4 \\
    mai\_Deva & 44.4 & 75.8 & 86.5 & 53.7 & 81.4 & 90.7 \\
    mal\_Mlym & 42.4 & 73.4 & 85.2 & 53.9 & 81.6 & 90.6 \\
    mar\_Deva & 43.9 & 74.5 & 85.1 & 53.7 & 79.9 & 89.9 \\
    min\_Latn & 40.2 & 70.8 & 82.7 & 50.4 & 77.5 & 86.0 \\
    mkd\_Cyrl & 44.8 & 74.3 & 85.8 & 56.7 & 80.5 & 90.2 \\
    mlt\_Latn & 41.8 & 71.7 & 83.7 & 52.7 & 78.3 & 88.5 \\
    mni\_Beng & 36.4 & 65.8 & 79.6 & 45.1 & 73.2 & 84.8 \\
    mos\_Latn & 23.5 & 47.6 & 59.5 & 29.2 & 54.0 & 63.2 \\
    mri\_Latn & 39.1 & 69.8 & 80.1 & 48.1 & 76.7 & 86.5 \\
    mya\_Mymr & 41.0 & 71.6 & 83.5 & 49.9 & 78.3 & 88.2 \\
    nld\_Latn & 46.3 & 76.5 & 86.9 & 56.5 & 83.0 & 91.9 \\
    nno\_Latn & 45.0 & 72.4 & 83.5 & 54.6 & 80.7 & 89.2 \\
    nob\_Latn & 46.0 & 75.0 & 86.0 & 56.7 & 82.3 & 91.2 \\
    npi\_Deva & 43.7 & 74.7 & 85.4 & 54.3 & 80.8 & 90.1 \\
    nso\_Latn & 42.2 & 73.3 & 83.0 & 53.8 & 79.2 & 88.5 \\
    nus\_Latn & 20.8 & 44.2 & 57.2 & 27.9 & 52.1 & 63.2 \\
    nya\_Latn & 42.2 & 71.8 & 83.1 & 50.5 & 77.9 & 88.4 \\
    oci\_Latn & 42.7 & 74.9 & 85.5 & 54.0 & 80.1 & 90.7 \\
    ory\_Orya & 42.4 & 75.1 & 85.7 & 53.2 & 80.2 & 89.7 \\
    pag\_Latn & 43.1 & 74.0 & 84.7 & 53.3 & 80.1 & 89.6 \\
    pan\_Guru & 43.4 & 74.2 & 85.4 & 56.2 & 81.1 & 89.7 \\
    pap\_Latn & 44.2 & 72.9 & 84.3 & 51.5 & 79.2 & 89.9 \\
    pbt\_Arab & 40.7 & 72.4 & 82.5 & 51.9 & 77.1 & 87.6 \\
    pes\_Arab & 44.8 & 74.0 & 85.0 & 55.4 & 80.7 & 90.1 \\
    plt\_Latn & 42.6 & 72.3 & 83.9 & 52.0 & 77.5 & 88.5 \\
    pol\_Latn & 44.9 & 75.4 & 85.2 & 56.4 & 82.4 & 91.4 \\
    por\_Latn & 46.7 & 75.9 & 86.6 & 58.3 & 81.7 & 92.0 \\
    prs\_Arab & 45.3 & 74.6 & 86.5 & 55.3 & 81.2 & 90.8 \\
    quy\_Latn & 28.0 & 54.5 & 65.4 & 34.5 & 61.0 & 71.5 \\
    ron\_Latn & 45.4 & 75.6 & 86.6 & 57.2 & 82.1 & 91.6 \\
    run\_Latn & 35.7 & 65.7 & 77.0 & 43.2 & 70.1 & 81.9 \\
    rus\_Cyrl & 45.4 & 75.6 & 86.3 & 55.7 & 82.3 & 91.9 \\
    sag\_Latn & 36.1 & 66.0 & 78.1 & 43.5 & 73.6 & 83.1 \\
    san\_Deva & 32.8 & 60.3 & 74.6 & 42.7 & 69.6 & 79.7 \\
    scn\_Latn & 43.8 & 73.5 & 84.7 & 55.2 & 80.3 & 90.3 \\
    shn\_Mymr & 28.0 & 54.3 & 66.4 & 31.9 & 59.2 & 72.3 \\
    sin\_Sinh & 41.3 & 72.5 & 84.0 & 51.8 & 78.1 & 87.5 \\
    slk\_Latn & 44.6 & 75.9 & 85.9 & 55.5 & 81.3 & 91.5 \\
    slv\_Latn & 45.4 & 75.6 & 85.2 & 56.2 & 80.2 & 90.2 \\
    smo\_Latn & 41.5 & 72.4 & 84.4 & 53.3 & 79.6 & 88.5 \\
    sna\_Latn & 41.1 & 71.7 & 83.1 & 51.3 & 78.8 & 89.3 \\
    snd\_Arab & 43.9 & 70.1 & 82.8 & 53.0 & 78.2 & 89.3 \\
    som\_Latn & 41.8 & 71.7 & 84.4 & 50.0 & 77.2 & 88.0 \\
    sot\_Latn & 42.6 & 74.7 & 84.2 & 54.8 & 80.1 & 88.7 \\
    spa\_Latn & 46.3 & 76.8 & 86.5 & 58.3 & 82.5 & 92.7 \\
    srd\_Latn & 41.8 & 67.7 & 80.7 & 50.8 & 74.7 & 86.1 \\
    srp\_Cyrl & 41.7 & 73.1 & 84.7 & 54.1 & 80.6 & 90.9 \\
    ssw\_Latn & 36.5 & 67.7 & 80.4 & 47.6 & 74.1 & 84.9 \\
    sun\_Latn & 45.1 & 74.6 & 86.1 & 56.0 & 81.8 & 89.9 \\
    swe\_Latn & 47.1 & 76.0 & 87.0 & 57.7 & 82.7 & 91.5 \\
    swh\_Latn & 44.0 & 73.0 & 84.0 & 53.5 & 80.3 & 89.3 \\
    szl\_Latn & 42.0 & 69.7 & 82.4 & 52.2 & 77.9 & 87.7 \\
    tam\_Taml & 42.9 & 72.2 & 83.9 & 53.7 & 79.4 & 89.1 \\
    taq\_Latn & 20.0 & 41.7 & 53.4 & 24.8 & 50.4 & 62.2 \\
    taq\_Tfng & 18.7 & 40.4 & 54.4 & 24.2 & 47.4 & 59.4 \\
    tat\_Cyrl & 41.0 & 71.4 & 84.0 & 51.2 & 78.8 & 89.2 \\
    tel\_Telu & 43.6 & 75.4 & 86.9 & 55.8 & 81.6 & 90.8 \\
    tgk\_Cyrl & 42.5 & 73.2 & 84.3 & 51.2 & 78.5 & 87.7 \\
    tgl\_Latn & 45.0 & 74.6 & 85.9 & 57.7 & 81.4 & 90.6 \\
    tha\_Thai & 42.2 & 71.7 & 81.9 & 50.1 & 78.0 & 87.9 \\
    tir\_Ethi & 39.0 & 68.9 & 81.6 & 47.1 & 75.1 & 86.7 \\
    tpi\_Latn & 40.1 & 70.9 & 82.9 & 47.7 & 79.1 & 88.2 \\
    tsn\_Latn & 41.7 & 71.2 & 83.8 & 53.3 & 77.8 & 88.5 \\
    tso\_Latn & 41.7 & 69.9 & 81.7 & 51.0 & 77.8 & 87.1 \\
    tuk\_Latn & 37.7 & 70.8 & 81.2 & 47.7 & 75.1 & 85.6 \\
    tum\_Latn & 34.8 & 63.6 & 75.2 & 43.2 & 69.8 & 79.9 \\
    tur\_Latn & 44.2 & 75.2 & 85.2 & 55.9 & 80.8 & 90.7 \\
    twi\_Latn & 35.0 & 62.4 & 74.4 & 41.8 & 69.3 & 79.9 \\
    tzm\_Tfng & 23.1 & 46.6 & 57.8 & 27.4 & 50.4 & 62.2 \\
    uig\_Arab & 38.6 & 68.3 & 80.1 & 46.5 & 73.9 & 83.9 \\
    ukr\_Cyrl & 42.9 & 74.6 & 86.2 & 54.9 & 83.3 & 92.2 \\
    umb\_Latn & 21.4 & 43.1 & 55.4 & 25.2 & 47.4 & 59.1 \\
    urd\_Arab & 42.8 & 74.3 & 85.6 & 56.1 & 81.4 & 90.6 \\
    uzn\_Latn & 42.1 & 72.0 & 83.2 & 52.0 & 77.8 & 88.6 \\
    vec\_Latn & 45.9 & 75.1 & 86.6 & 57.1 & 81.9 & 91.5 \\
    vie\_Latn & 45.1 & 74.9 & 85.3 & 56.1 & 80.3 & 90.2 \\
    war\_Latn & 40.5 & 69.8 & 81.7 & 52.0 & 76.0 & 86.0 \\
    wol\_Latn & 27.1 & 54.4 & 67.8 & 36.7 & 62.1 & 73.7 \\
    xho\_Latn & 41.7 & 72.1 & 83.3 & 52.2 & 79.0 & 87.0 \\
    ydd\_Hebr & 42.8 & 72.6 & 84.6 & 52.8 & 79.9 & 90.7 \\
    yor\_Latn & 38.3 & 68.4 & 80.7 & 48.6 & 72.9 & 86.1 \\
    yue\_Hant & 41.4 & 71.1 & 83.1 & 51.1 & 78.0 & 87.8 \\
    zho\_Hans & 42.0 & 72.3 & 84.0 & 50.6 & 79.6 & 89.3 \\
    zho\_Hant & 37.4 & 66.5 & 79.8 & 47.2 & 73.0 & 84.9 \\
    zsm\_Latn & 44.9 & 75.7 & 84.7 & 55.1 & 82.2 & 90.4 \\
    zul\_Latn & 42.1 & 72.1 & 83.7 & 52.4 & 80.3 & 89.4 \\
    \midrule
    Average & 39.4 & 68.6 & 80.0 & 49.1 & 75.2 & 85.3 \\
\end{longtable}

\begin{longtable}{lcccccc}
    \caption{Flickr30k-200 evaluation dataset results} 
    \label{tab:flickr30k_200} \\
    \toprule
     & \multicolumn{3}{c}{NLLB-CLIP base} & \multicolumn{3}{c}{NLLB-CLIP large} \\
    \cmidrule(lr){2-4}
    \cmidrule(lr){5-7}
    Language & R@1 & R@5 & R@10 & R@1 & R@5 & R@10 \\
    \midrule
    \endfirsthead 

    \multicolumn{7}{c}{{\bfseries \tablename\ \thetable{} -- continued from the previous page}} \\
    \midrule
    & \multicolumn{3}{c}{NLLB-CLIP base} & \multicolumn{3}{c}{NLLB-CLIP large} \\
    \cmidrule(lr){2-4}
    \cmidrule(lr){5-7}
    Language & R@1 & R@5 & R@10 & R@1 & R@5 & R@10 \\
    \midrule
    \endhead 

    \midrule \multicolumn{7}{r}{{Continued on the next page}} \\
    \endfoot 

    \bottomrule
    \endlastfoot 

    ace\_Arab & 20.1 & 41.3 & 50.3 & 28.0 & 51.3 & 58.8 \\
    ace\_Latn & 42.4 & 68.3 & 78.6 & 52.3 & 77.5 & 86.0 \\
    acm\_Arab & 48.4 & 74.8 & 82.0 & 61.2 & 83.9 & 88.9 \\
    acq\_Arab & 50.0 & 76.7 & 84.0 & 62.7 & 85.7 & 91.3 \\
    aeb\_Arab & 38.7 & 63.3 & 72.7 & 47.6 & 71.5 & 78.9 \\
    afr\_Latn & 51.0 & 78.2 & 86.8 & 64.2 & 86.8 & 92.8 \\
    ajp\_Arab & 49.2 & 76.1 & 84.1 & 61.4 & 83.9 & 90.0 \\
    aka\_Latn & 39.2 & 66.6 & 76.1 & 49.1 & 76.3 & 84.1 \\
    als\_Latn & 50.2 & 76.3 & 84.2 & 60.2 & 86.7 & 91.5 \\
    amh\_Ethi & 47.6 & 73.7 & 80.8 & 58.6 & 83.8 & 89.2 \\
    apc\_Arab & 49.4 & 75.6 & 83.5 & 61.6 & 84.9 & 89.6 \\
    arb\_Arab & 51.2 & 78.0 & 84.7 & 63.7 & 87.2 & 91.6 \\
    ars\_Arab & 49.8 & 76.7 & 84.9 & 62.6 & 85.5 & 91.9 \\
    ary\_Arab & 45.1 & 70.4 & 79.1 & 57.6 & 78.6 & 85.6 \\
    arz\_Arab & 49.7 & 76.7 & 83.9 & 63.2 & 86.1 & 91.6 \\
    asm\_Beng & 45.5 & 74.0 & 82.7 & 58.8 & 82.9 & 88.6 \\
    ast\_Latn & 50.6 & 75.7 & 83.5 & 63.2 & 85.6 & 90.4 \\
    awa\_Deva & 38.4 & 64.0 & 71.8 & 49.8 & 70.8 & 78.1 \\
    ayr\_Latn & 39.4 & 67.7 & 76.6 & 51.1 & 75.1 & 83.1 \\
    azb\_Arab & 43.3 & 70.0 & 79.2 & 53.4 & 78.9 & 86.3 \\
    azj\_Latn & 51.3 & 77.5 & 84.3 & 62.9 & 87.9 & 92.2 \\
    bak\_Cyrl & 47.6 & 73.8 & 82.3 & 58.4 & 83.5 & 90.3 \\
    bam\_Latn & 33.3 & 62.1 & 72.4 & 43.5 & 71.1 & 80.6 \\
    ban\_Latn & 40.6 & 69.1 & 78.5 & 51.5 & 77.8 & 85.7 \\
    bel\_Cyrl & 49.9 & 77.8 & 84.8 & 61.4 & 85.3 & 90.4 \\
    bem\_Latn & 37.6 & 64.2 & 73.2 & 48.3 & 72.2 & 81.9 \\
    ben\_Beng & 51.1 & 78.9 & 84.9 & 63.2 & 87.8 & 91.0 \\
    bho\_Deva & 49.4 & 76.9 & 83.4 & 61.6 & 84.5 & 90.0 \\
    bjn\_Arab & 18.7 & 37.7 & 47.0 & 22.2 & 43.0 & 52.4 \\
    bjn\_Latn & 47.6 & 74.9 & 84.4 & 60.0 & 84.5 & 90.0 \\
    bod\_Tibt & 35.6 & 63.2 & 74.0 & 45.8 & 71.1 & 79.6 \\
    bos\_Latn & 51.5 & 77.4 & 85.0 & 63.8 & 86.3 & 91.9 \\
    bug\_Latn & 40.1 & 66.9 & 76.4 & 49.8 & 74.5 & 83.0 \\
    bul\_Cyrl & 51.4 & 77.1 & 85.2 & 63.8 & 86.6 & 92.5 \\
    cat\_Latn & 50.5 & 78.7 & 84.5 & 64.3 & 86.9 & 92.7 \\
    ceb\_Latn & 51.4 & 77.9 & 85.3 & 63.1 & 85.5 & 91.2 \\
    ces\_Latn & 50.4 & 78.1 & 85.5 & 64.7 & 87.2 & 91.3 \\
    cjk\_Latn & 26.3 & 50.4 & 60.5 & 31.4 & 58.1 & 69.0 \\
    ckb\_Arab & 47.8 & 73.9 & 83.0 & 58.3 & 81.7 & 88.7 \\
    crh\_Latn & 49.0 & 75.4 & 83.6 & 62.1 & 85.6 & 91.3 \\
    cym\_Latn & 46.2 & 72.4 & 81.5 & 57.3 & 82.7 & 90.4 \\
    dan\_Latn & 53.5 & 79.1 & 85.9 & 63.9 & 87.7 & 92.9 \\
    deu\_Latn & 53.0 & 80.1 & 86.5 & 63.0 & 88.9 & 92.7 \\
    dik\_Latn & 26.7 & 52.4 & 62.1 & 33.4 & 57.9 & 67.4 \\
    dyu\_Latn & 20.2 & 41.6 & 52.8 & 25.8 & 48.5 & 59.1 \\
    dzo\_Tibt & 31.9 & 59.4 & 68.3 & 41.1 & 68.6 & 78.0 \\
    ell\_Grek & 51.4 & 78.8 & 84.3 & 63.6 & 87.9 & 92.2 \\
    epo\_Latn & 52.3 & 77.4 & 85.1 & 64.7 & 86.8 & 92.5 \\
    est\_Latn & 48.2 & 74.8 & 83.3 & 59.9 & 84.9 & 89.6 \\
    eus\_Latn & 46.1 & 75.2 & 81.6 & 58.1 & 82.1 & 89.8 \\
    ewe\_Latn & 40.0 & 68.1 & 77.0 & 48.4 & 75.6 & 84.4 \\
    fao\_Latn & 48.1 & 73.7 & 82.4 & 58.5 & 83.8 & 89.6 \\
    fij\_Latn & 41.9 & 69.9 & 78.4 & 55.0 & 80.4 & 86.9 \\
    fin\_Latn & 49.1 & 76.2 & 83.0 & 61.3 & 84.8 & 90.8 \\
    fon\_Latn & 29.7 & 53.3 & 63.0 & 37.7 & 63.3 & 73.3 \\
    fra\_Latn & 53.2 & 77.3 & 87.2 & 63.9 & 88.6 & 93.8 \\
    fur\_Latn & 46.3 & 72.9 & 81.0 & 59.5 & 83.1 & 90.2 \\
    fuv\_Latn & 31.1 & 58.7 & 68.2 & 38.3 & 63.6 & 74.5 \\
    gaz\_Latn & 37.3 & 65.1 & 74.8 & 50.7 & 77.0 & 84.6 \\
    gla\_Latn & 47.3 & 73.3 & 81.6 & 57.9 & 83.3 & 89.6 \\
    gle\_Latn & 44.3 & 73.4 & 81.3 & 58.7 & 83.9 & 89.9 \\
    glg\_Latn & 51.1 & 79.0 & 85.9 & 64.6 & 88.2 & 92.6 \\
    grn\_Latn & 44.7 & 69.3 & 79.0 & 54.9 & 79.3 & 85.8 \\
    guj\_Gujr & 51.8 & 76.5 & 84.7 & 64.1 & 86.4 & 91.8 \\
    hat\_Latn & 51.1 & 76.3 & 84.7 & 61.6 & 85.9 & 91.1 \\
    hau\_Latn & 46.8 & 72.1 & 81.2 & 58.0 & 82.3 & 88.7 \\
    heb\_Hebr & 47.6 & 75.6 & 84.9 & 62.3 & 85.5 & 91.9 \\
    hin\_Deva & 53.1 & 79.1 & 85.7 & 64.3 & 87.3 & 91.7 \\
    hne\_Deva & 51.3 & 78.3 & 83.8 & 63.7 & 85.6 & 90.8 \\
    hrv\_Latn & 50.4 & 77.5 & 85.0 & 64.3 & 86.0 & 91.3 \\
    hun\_Latn & 50.0 & 76.8 & 86.2 & 63.2 & 86.6 & 91.7 \\
    hye\_Armn & 47.8 & 74.9 & 83.3 & 59.0 & 83.1 & 89.2 \\
    ibo\_Latn & 44.0 & 73.9 & 82.3 & 55.3 & 82.9 & 88.7 \\
    ilo\_Latn & 48.7 & 73.7 & 83.5 & 59.1 & 83.8 & 88.5 \\
    ind\_Latn & 51.4 & 79.9 & 86.4 & 65.6 & 87.5 & 91.9 \\
    isl\_Latn & 46.8 & 74.5 & 83.1 & 58.2 & 83.8 & 90.1 \\
    ita\_Latn & 50.7 & 78.8 & 85.9 & 65.6 & 87.6 & 92.2 \\
    jav\_Latn & 51.1 & 77.6 & 86.2 & 63.0 & 85.6 & 91.2 \\
    jpn\_Jpan & 48.5 & 73.7 & 82.9 & 60.2 & 80.9 & 89.0 \\
    kab\_Latn & 35.4 & 62.2 & 72.5 & 44.6 & 72.2 & 81.7 \\
    kac\_Latn & 33.6 & 62.5 & 71.7 & 41.8 & 69.6 & 77.3 \\
    kam\_Latn & 10.5 & 24.2 & 32.9 & 12.0 & 28.4 & 37.4 \\
    kan\_Knda & 50.4 & 77.8 & 84.1 & 62.2 & 86.3 & 91.5 \\
    kas\_Arab & 38.4 & 64.4 & 71.6 & 48.4 & 75.3 & 81.6 \\
    kas\_Deva & 42.2 & 67.5 & 76.5 & 52.4 & 76.8 & 84.1 \\
    kat\_Geor & 43.1 & 69.9 & 78.3 & 54.6 & 78.5 & 85.5 \\
    kaz\_Cyrl & 48.3 & 74.4 & 83.1 & 58.4 & 84.0 & 89.0 \\
    kbp\_Latn & 25.5 & 48.1 & 58.2 & 29.8 & 55.1 & 64.6 \\
    kea\_Latn & 44.7 & 71.4 & 79.5 & 55.6 & 81.5 & 88.7 \\
    khk\_Cyrl & 37.6 & 65.1 & 76.6 & 48.5 & 77.1 & 84.8 \\
    khm\_Khmr & 39.9 & 66.7 & 77.2 & 51.6 & 76.3 & 84.8 \\
    kik\_Latn & 38.2 & 66.1 & 74.8 & 47.1 & 73.5 & 82.0 \\
    kin\_Latn & 44.3 & 72.5 & 80.9 & 56.2 & 81.1 & 88.7 \\
    kir\_Cyrl & 45.6 & 71.7 & 80.5 & 56.9 & 82.0 & 88.0 \\
    kmb\_Latn & 22.0 & 44.4 & 55.6 & 26.8 & 51.8 & 61.0 \\
    kmr\_Latn & 40.9 & 70.1 & 79.7 & 52.5 & 79.4 & 86.5 \\
    knc\_Arab & 37.4 & 64.7 & 74.9 & 49.7 & 74.6 & 82.7 \\
    knc\_Latn & 23.2 & 47.6 & 57.2 & 30.6 & 52.7 & 63.5 \\
    kon\_Latn & 43.4 & 69.3 & 79.4 & 52.9 & 79.0 & 86.3 \\
    kor\_Hang & 51.0 & 75.8 & 84.6 & 61.8 & 85.5 & 91.0 \\
    lao\_Laoo & 45.9 & 72.9 & 82.1 & 56.8 & 83.1 & 89.4 \\
    lij\_Latn & 45.7 & 71.5 & 80.7 & 57.8 & 81.9 & 87.6 \\
    lim\_Latn & 48.3 & 76.6 & 84.5 & 61.1 & 85.8 & 91.3 \\
    lin\_Latn & 45.6 & 73.4 & 82.1 & 58.2 & 81.7 & 89.7 \\
    lit\_Latn & 48.6 & 74.7 & 82.5 & 58.4 & 84.1 & 90.5 \\
    lmo\_Latn & 47.9 & 75.1 & 82.6 & 60.1 & 84.2 & 90.4 \\
    ltg\_Latn & 38.7 & 65.7 & 74.9 & 48.9 & 76.0 & 84.3 \\
    ltz\_Latn & 51.0 & 77.2 & 84.5 & 63.3 & 86.5 & 91.8 \\
    lua\_Latn & 37.2 & 66.4 & 76.1 & 46.9 & 76.2 & 84.2 \\
    lug\_Latn & 39.8 & 66.6 & 78.0 & 52.0 & 78.1 & 86.2 \\
    luo\_Latn & 39.5 & 66.8 & 75.4 & 50.7 & 77.4 & 85.0 \\
    lus\_Latn & 37.4 & 62.7 & 73.1 & 45.3 & 69.7 & 78.7 \\
    lvs\_Latn & 44.8 & 71.9 & 80.5 & 57.3 & 81.5 & 88.2 \\
    mag\_Deva & 52.1 & 78.4 & 85.0 & 63.0 & 85.8 & 91.8 \\
    mai\_Deva & 51.1 & 76.4 & 85.0 & 62.3 & 86.7 & 91.5 \\
    mal\_Mlym & 49.2 & 75.7 & 83.3 & 59.7 & 85.0 & 90.2 \\
    mar\_Deva & 50.9 & 77.6 & 84.8 & 62.5 & 85.5 & 91.0 \\
    min\_Latn & 47.4 & 74.0 & 82.7 & 58.0 & 83.3 & 88.2 \\
    mkd\_Cyrl & 51.4 & 76.4 & 83.4 & 61.5 & 85.9 & 91.6 \\
    mlt\_Latn & 46.7 & 72.7 & 81.5 & 59.2 & 84.5 & 90.1 \\
    mni\_Beng & 38.6 & 63.8 & 72.3 & 49.0 & 74.3 & 81.6 \\
    mos\_Latn & 30.3 & 55.2 & 67.1 & 38.6 & 64.5 & 73.2 \\
    mri\_Latn & 42.6 & 72.2 & 81.2 & 55.4 & 82.2 & 88.3 \\
    mya\_Mymr & 45.6 & 70.3 & 79.3 & 55.5 & 80.1 & 84.6 \\
    nld\_Latn & 51.0 & 77.7 & 85.7 & 64.1 & 88.1 & 92.6 \\
    nno\_Latn & 51.4 & 77.1 & 84.8 & 61.2 & 85.2 & 90.8 \\
    nob\_Latn & 52.2 & 78.6 & 85.8 & 63.6 & 87.8 & 92.3 \\
    npi\_Deva & 49.0 & 74.8 & 82.0 & 60.2 & 83.2 & 88.6 \\
    nso\_Latn & 48.8 & 76.2 & 84.3 & 60.1 & 84.2 & 90.0 \\
    nus\_Latn & 26.0 & 48.3 & 58.6 & 32.9 & 56.8 & 67.8 \\
    nya\_Latn & 48.3 & 73.5 & 83.1 & 60.0 & 83.4 & 89.5 \\
    oci\_Latn & 50.8 & 76.7 & 84.2 & 64.1 & 86.4 & 91.7 \\
    ory\_Orya & 47.4 & 73.9 & 83.0 & 59.6 & 84.1 & 90.2 \\
    pag\_Latn & 48.0 & 74.8 & 82.9 & 57.7 & 83.3 & 89.2 \\
    pan\_Guru & 51.2 & 77.3 & 84.3 & 63.3 & 86.2 & 91.2 \\
    pap\_Latn & 49.9 & 75.7 & 83.9 & 61.0 & 85.9 & 90.7 \\
    pbt\_Arab & 49.3 & 73.7 & 81.8 & 61.4 & 83.7 & 89.7 \\
    pes\_Arab & 50.8 & 77.4 & 84.5 & 63.7 & 86.3 & 91.5 \\
    plt\_Latn & 48.3 & 73.8 & 83.7 & 60.6 & 84.6 & 90.4 \\
    pol\_Latn & 49.9 & 76.6 & 84.4 & 63.0 & 86.4 & 91.8 \\
    por\_Latn & 52.7 & 79.1 & 86.6 & 65.4 & 88.8 & 92.5 \\
    prs\_Arab & 50.5 & 77.0 & 84.4 & 65.0 & 87.0 & 92.1 \\
    quy\_Latn & 31.9 & 58.6 & 68.2 & 40.0 & 66.3 & 74.7 \\
    ron\_Latn & 51.2 & 77.1 & 84.7 & 64.5 & 87.1 & 92.2 \\
    run\_Latn & 43.0 & 71.6 & 80.3 & 56.3 & 80.6 & 87.5 \\
    rus\_Cyrl & 51.0 & 77.9 & 84.5 & 62.6 & 87.0 & 92.3 \\
    sag\_Latn & 41.0 & 68.7 & 76.7 & 52.1 & 77.2 & 86.1 \\
    san\_Deva & 38.4 & 63.2 & 73.2 & 48.2 & 73.1 & 80.9 \\
    scn\_Latn & 49.6 & 75.9 & 84.0 & 62.7 & 85.9 & 91.7 \\
    shn\_Mymr & 32.9 & 60.3 & 72.3 & 43.5 & 70.4 & 78.8 \\
    sin\_Sinh & 47.0 & 72.2 & 80.1 & 58.2 & 81.8 & 86.9 \\
    slk\_Latn & 49.2 & 77.2 & 85.6 & 62.4 & 87.1 & 91.7 \\
    slv\_Latn & 48.7 & 76.5 & 84.0 & 61.7 & 85.8 & 91.2 \\
    smo\_Latn & 48.1 & 73.3 & 82.3 & 58.9 & 83.7 & 90.4 \\
    sna\_Latn & 47.4 & 73.3 & 82.8 & 60.4 & 84.8 & 91.7 \\
    snd\_Arab & 47.0 & 73.9 & 82.2 & 60.4 & 84.0 & 89.5 \\
    som\_Latn & 46.8 & 75.5 & 83.1 & 60.2 & 84.5 & 88.8 \\
    sot\_Latn & 48.8 & 76.5 & 83.8 & 61.7 & 85.4 & 91.7 \\
    spa\_Latn & 51.8 & 79.4 & 85.7 & 65.4 & 88.7 & 93.4 \\
    srd\_Latn & 47.5 & 73.6 & 82.6 & 59.3 & 83.4 & 90.7 \\
    srp\_Cyrl & 50.4 & 75.6 & 83.8 & 61.7 & 84.9 & 90.7 \\
    ssw\_Latn & 41.9 & 69.1 & 79.8 & 53.6 & 80.1 & 87.2 \\
    sun\_Latn & 50.3 & 77.3 & 85.2 & 63.1 & 86.7 & 92.3 \\
    swe\_Latn & 52.9 & 78.2 & 85.5 & 62.8 & 86.7 & 91.9 \\
    swh\_Latn & 51.4 & 76.0 & 84.5 & 61.6 & 85.6 & 90.8 \\
    szl\_Latn & 45.4 & 72.3 & 81.6 & 58.2 & 82.9 & 89.9 \\
    tam\_Taml & 49.6 & 75.3 & 83.1 & 59.8 & 84.4 & 89.8 \\
    taq\_Latn & 20.7 & 40.7 & 49.9 & 24.7 & 46.2 & 55.3 \\
    taq\_Tfng & 14.9 & 35.0 & 44.8 & 21.0 & 41.9 & 53.5 \\
    tat\_Cyrl & 49.2 & 75.3 & 83.2 & 60.0 & 84.4 & 90.6 \\
    tel\_Telu & 48.9 & 77.3 & 85.3 & 61.9 & 86.5 & 90.7 \\
    tgk\_Cyrl & 46.0 & 74.3 & 82.6 & 58.4 & 82.3 & 88.0 \\
    tgl\_Latn & 51.4 & 77.2 & 86.3 & 63.5 & 86.9 & 92.4 \\
    tha\_Thai & 48.0 & 75.3 & 83.8 & 60.0 & 83.7 & 89.9 \\
    tir\_Ethi & 43.2 & 71.6 & 79.2 & 55.6 & 80.6 & 87.2 \\
    tpi\_Latn & 46.6 & 72.3 & 81.0 & 57.2 & 81.5 & 88.1 \\
    tsn\_Latn & 49.9 & 75.5 & 84.0 & 61.5 & 85.3 & 90.7 \\
    tso\_Latn & 47.0 & 74.7 & 83.7 & 59.3 & 83.8 & 90.9 \\
    tuk\_Latn & 46.0 & 71.1 & 81.3 & 55.0 & 81.8 & 87.9 \\
    tum\_Latn & 39.5 & 67.7 & 76.9 & 47.8 & 75.0 & 83.5 \\
    tur\_Latn & 52.6 & 78.3 & 85.6 & 64.8 & 87.1 & 91.7 \\
    twi\_Latn & 38.8 & 67.2 & 76.5 & 48.9 & 75.4 & 83.8 \\
    tzm\_Tfng & 23.2 & 46.4 & 56.6 & 30.2 & 55.1 & 64.5 \\
    uig\_Arab & 43.0 & 69.7 & 78.7 & 54.2 & 79.9 & 86.1 \\
    ukr\_Cyrl & 52.5 & 77.6 & 84.9 & 62.8 & 86.0 & 91.5 \\
    umb\_Latn & 19.7 & 41.4 & 53.4 & 24.2 & 48.3 & 60.2 \\
    urd\_Arab & 51.5 & 76.1 & 84.9 & 63.5 & 85.9 & 92.1 \\
    uzn\_Latn & 47.4 & 76.1 & 83.2 & 58.5 & 84.6 & 89.9 \\
    vec\_Latn & 49.9 & 77.9 & 85.3 & 64.4 & 86.1 & 91.8 \\
    vie\_Latn & 50.5 & 77.7 & 85.9 & 62.2 & 86.2 & 92.0 \\
    war\_Latn & 48.7 & 74.3 & 82.1 & 58.9 & 82.4 & 87.6 \\
    wol\_Latn & 30.0 & 57.5 & 67.9 & 40.4 & 68.4 & 77.7 \\
    xho\_Latn & 48.5 & 74.1 & 83.9 & 61.8 & 84.9 & 90.4 \\
    ydd\_Hebr & 47.0 & 74.7 & 83.8 & 60.2 & 84.0 & 90.5 \\
    yor\_Latn & 42.7 & 70.7 & 79.6 & 54.1 & 80.5 & 87.7 \\
    yue\_Hant & 48.0 & 74.5 & 83.5 & 57.0 & 81.2 & 87.4 \\
    zho\_Hans & 47.9 & 74.5 & 81.4 & 58.0 & 83.0 & 88.3 \\
    zho\_Hant & 30.6 & 54.4 & 63.1 & 37.6 & 60.7 & 69.1 \\
    zsm\_Latn & 51.6 & 78.2 & 86.2 & 63.8 & 87.5 & 91.2 \\
    zul\_Latn & 47.9 & 75.2 & 83.9 & 61.3 & 85.3 & 91.4 \\
    \midrule
    Average & 44.49 & 70.81 & 79.32 & 55.56 & 79.86 & 86.37 \\
\end{longtable}

\end{document}